\documentclass[letterpaper, 10 pt, conference]{IEEEtran}

\IEEEoverridecommandlockouts %

\usepackage{subfig}

\usepackage[style=ieee, backend=biber, bibencoding=utf8, maxbibnames=3, mincitenames=1, maxcitenames=2, url=false, doi=false, isbn=false, citestyle=numeric-comp, natbib=true, eprint=false]{biblatex}

\usepackage{graphicx} 
\usepackage{bmpsize}
\graphicspath{{figures/}} %

\usepackage{scrhack} %
\usepackage{listings}
\usepackage{lstautogobble}
\usepackage{tikz}
\usepackage{booktabs} %

\usepackage[utf8]{inputenc} %

\usetikzlibrary{external}

\usepackage{pgfplots}
\usepackage{pgfplotstable}

\usepackage{csquotes}

\bibliography{bib/library}

\usepackage[colorinlistoftodos,prependcaption,textsize=small]{todonotes}

\usepackage{bibentry}

\usepackage{siunitx}
\usepackage{xargs} %

\usepackage{amsmath}

\usepackage{scalerel}

\usepackage{amssymb}

\usepackage{caption}
\usepackage{tikz}
\usetikzlibrary{shapes, arrows, backgrounds, positioning, fit, chains}

\usepackage{amsthm}
\usepackage{thmtools}

\declaretheoremstyle[
spaceabove={},%
spacebelow={},%
headfont=\normalfont\bfseries,
notefont=\bfseries, 
notebraces={}{},
bodyfont=\normalfont,
postheadspace=0.5em,
name={\ignorespaces},
numbered=no,
headpunct=:]
{rulecitationstyle}
\declaretheorem[style=rulecitationstyle,shaded={bgcolor=white}]{trafficrule}
\newcommand{\refFigure}[1]{Fig. \ref{#1}}

\newcommand{\refSection}[1]{Section \ref{#1}}

\newcommand{\refTable}[1]{Table\space\ref{#1}}

\newcommand{\clearemptydoublepage}{%
  \ifthenelse{\boolean{@twoside}}{\newpage{\pagestyle{empty}\cleardoublepage}}%
  {\clearpage}}

\newcolumntype{L}[1]{>{\raggedright\arraybackslash}p{#1}}

\newcolumntype{C}[1]{>{\centering\arraybackslash}p{#1}}

\newcolumntype{R}[1]{>{\raggedleft\arraybackslash}p{#1}}

\newcommandx{\thought}[2][1=]{\todo[inline,linecolor=blue,backgroundcolor=green!25,bordercolor=green,#1]{#2}}
\newcommandx{\legalanalysis}[2][1=]{\todo[inline,linecolor=yellow,backgroundcolor=yellow!25,bordercolor=yellow,#1]{#2}}
\newcommandx{\change}[2][1=]{\todo[inline,linecolor=blue,backgroundcolor=blue!25,bordercolor=blue,#1]{#2}}
\newcommand \Global {\scaleobj{0.95}{\square}}
\newcommand \Next       {\scaleobj{0.85}{\bigcirc}}
\newcommand \Final       {\lozenge}
\newcommand \lUntil      {\mathsf{U}} %
\newcommand \ImpliesCustomized {\scaleobj{0.9}{\implies}}

\newcommand \AtomicProposition {\pi}
\newcommand \AtomicPropositionSet {\Pi}

\newcommand \LTLformula {\varphi}

\newcommand \true {\mathbf{T}}

\newcommand{\labelstyle}[1]{\footnotesize{\textsf{\textit{{#1}}}}}
\newcommand \labelInDirectFront {\labelstyle{pred}}
\newcommand \labelRight {\labelstyle{right}}
\newcommand \labelLeft {\labelstyle{left}}
\newcommand \labelFront {\labelstyle{in-front}}
\newcommand \labelBehind {\labelstyle{behind}}
\newcommand \labelMerged {\labelstyle{merged}}
\newcommand \labelSafeDistanceFront {\labelstyle{sd-front}}
\newcommand \labelSafeDistanceRear {\labelstyle{sd-rear}}
\newcommand \labelColliding {\labelstyle{collide}}
\newcommand \labelLaneChange {\labelstyle{lane-change}}
\newcommand \labelNear {\labelstyle{near}}
\newcommand \labelLaneEnd {\labelstyle{lane-end}}
\newcommand \labelAccelerate {\labelstyle{acc}}
\newcommand \labelSpeedDiff {\labelstyle{speed-adv}}
\newcommand \labelDivergingLane {\labelstyle{div-lane}}
\newcommand \labelAccelerationLane {\labelstyle{acc-lane}}
\newcommand \labelBuiltup {\labelstyle{built-up}}
\newcommand \labelMotorway {\labelstyle{motorway}}
\newcommand \labelDense {\labelstyle{dense}}

\newcommand{\labelEgoAgent}{i}
\newcommand{\labelInRelationToAgent}{j}
\newcommand{\labelInRelationToSecondAgent}{k}
\newcommand{\inrelationto}{\rightarrowtail}
\newcommand{\itoj}{^{(\labelEgoAgent \inrelationto \labelInRelationToAgent)}}
\newcommand{\ionly}{^{(\labelEgoAgent)}}

\newcommand {\AccLimit} {a_{\mathrm{lim}}}
\newcommand {\NearDistance} {d_{\mathrm{near}}}
\newcommand {\RemainingDistance} {s_{\mathrm{rem}}}
\newcommand {\VelocityThreshold} {v_{\mathrm{diff}}}
\newcommand {\DenseRadius} {r_{\mathrm{dense}}}
\newcommand {\DenseNumAgents} {N_{\mathrm{dense}}}

\newcommand{\Agenti}[1][]{#1}

\newcommand{\classnamestyle}[1]{\small{\textsf{{#1}}}}

\newcommand{\techreal}{\text{IR}}
\newcommand{\techltl}{\text{LTL}}
\newcommand{\techmissing}{\text{X}}

\tikzstyle{assumption} = [rectangle, draw, fill=gray!20, text width=11em, text centered, minimum height=2em]
\tikzstyle{assumption-except} = [rectangle, draw=red, fill=gray!20,  text width=11em, text centered, minimum height=2em]
\tikzstyle{guarantee} = [rectangle, draw, fill=gray!20, text width=11em, text centered, minimum height=2em]
\tikzstyle{line} = [draw, -latex']
\tikzstyle{line-except} = [draw, color=red, -latex']
\tikzstyle{premise-conclusion} = [draw, dashed, line width = 0.4pt, rounded corners=.1cm, inner sep=.2cm, -latex']

\newcommand*\playersinv[1]{({#1})}
\title{\LARGE \bf Formalizing Traffic Rules for Machine Interpretability}
\author{Klemens Esterle$^{1}$, Luis Gressenbuch$^{1}$ and Alois Knoll$^{2}$%
	\thanks{$^{1}$fortiss GmbH, Research Institute of the Free State of Bavaria, Munich, Germany}%
	\thanks{$^{2}$Alois Knoll is with Robotics, Artificial Intelligence and Real-time Systems, Technische Universit\"{a}t M\"{u}nchen, Munich, Germany}%
}

\newcommand\copyrighttext{%
	\scriptsize \textcolor{blue}{\textcopyright 2020 IEEE. Personal use of this material is permitted.  Permission from IEEE must be obtained for all other uses, in any current or future media, including reprinting/republishing this material for advertising or promotional purposes, creating new collective works, for resale or redistribution to servers or lists, or reuse of any copyrighted component of this work in other works}}
\newcommand\copyrightnotice{%
	\begin{tikzpicture}[remember picture,overlay]
	\node[anchor=north,yshift=-7.5pt] at (current page.north) {\fbox{\parbox{\dimexpr\textwidth-\fboxsep-\fboxrule\relax}{\copyrighttext}}};
	\end{tikzpicture}%
}

\begin{document}

\maketitle
\copyrightnotice
\thispagestyle{empty}
\pagestyle{empty}

\global\csname @topnum\endcsname 0
\global\csname @botnum\endcsname 0

\begin{abstract}
Autonomous vehicles need to be designed to abide by the same rules that humans follow. This is challenging, because traffic rules are fuzzy and not well defined, making them incomprehensible to machines.
Satisfaction cannot be incorporated in a planning component without proper formalization, nor can it be monitored and verified during simulation or testing. However, no research work has provided a consistent set of machine-interpretable traffic rules for a given operational driving domain.
In this paper, we propose a methodology for the legal study and formalization of traffic rules in a formal language. We use  Linear Temporal Logic as a formal specification language to describe temporal behaviors, capable of capturing a wide range of traffic rules.
We contribute a formalized set of traffic rules for dual carriageways and evaluate the effectiveness of our formalized rules on a public dataset.

\end{abstract}

\IEEEpeerreviewmaketitle

\section{Introduction}
\label{sec:introduction}
Traffic regulations such as the \textit{Straßenverkehrsordnung} (StVO) \cite{BMJV2013}, which is the German concretization of the Vienna Convention on Road Traffic \cite{EconomicCommissionforEuropeInlandTransportCommittee1968}, define rules all drivers should obey.
These traffic rules are often fuzzy and subject to interpretation, encouraging the need for a formalized machine-interpretable definition of traffic rules. 
This formalization is essential for the development of a planning component, i.e. vehicles will adhere to the rules at all times. It may also support simulation-based verification or case-law.

The traditional approach in the planning community has been to represent legal aspects such as speed limits or traffic lights as geometric obstacles in space-time \cite{Ajanovic2018}, often forming spatio-temporal driving corridors that vehicles are allowed to operate. However, while such an approach works for static rules which can be easily mapped to constraints, it does not scale to more complex behavioral rules with multiple agents.
To formalize legal aspects, natural language must be translated. Logical languages are a formal way to represent rules. A logical language needs (1) to be expressive enough to codify natural language and (2) to have a mechanism for model-checking the formulas (i.e. traffic rules). 
Previous works have identified Linear Temporal Logic (LTL) as a suitable formal language to specify traffic rules \cite{ReyesCastro2013,  Rizaldi2015, Esterle2019a}.
Other works have used inequality constraints based on real numbers to formalize traffic rules \cite{Vanholme2013, Decastro2018}.

However, no work has yet provided a valid and consistent set of traffic rules for a restricted operational area. Likewise, there is no methodology to derive such rules.
The contributions of this paper are a methodology to formalize traffic rules from legal texts to a formal language, and a formalized set of traffic rules for dual carriageways. We evaluate these rules on a public dataset, which helped us to identify errors in the predicates but eventually provides valuable insight into the extent to which humans follow these rules.

\section{Legal Analysis of German Traffic Rules on Dual Carriageways in On-Ramp Scenarios}
At first, the operational design domain needs to be defined. We will analyze traffic rules for a passenger vehicle\footnote{The vehicle does not exceed 3.5 tons, is legally allowed to drive on motorways, and has no trailer.} based on the StVO\footnote{An English translation of the StVO is available at \\\url{https://germanlawarchive.iuscomp.org/}}, while also including references to the Vienna Convention on Road Traffic \cite{EconomicCommissionforEuropeInlandTransportCommittee1968}. 
Specifically, we analyze rules applicable to dual carriageways, such as highways. We do not consider rules that include pedestrians or cyclists.

Our interest is mainly focused on behavioral rules for road users, especially for multiple road users involved. We will not consider the following special cases:
\begin{itemize}
	\item Parking, breakdowns, and towing
	\item Necessary post-accident actions including clearing
	\item Signaling such as indicator signals or lighting 
	\item Regulatory signs, including lane markings, informatory signs, and traffic installations
	\item Limited visibility
\end{itemize}

To identify all relevant rules and remove ambiguity in them, we will use the following methodology:
\begin{enumerate}
	\item Identify a rule and separate it into an initial premise and a conclusion. If there is no premise, start with \blockquote{always}.
	\item Identify all exceptions to the premise. Use negated exceptions to update the premise.
	\item Decompose the premise and conclusion into labels or new rules.
\end{enumerate}
Using Boolean laws, we can combine the initial premise and exceptions by conjunction. We will use a graphical representation for identification and aggregation. Implications are illustrated by an arrow. Exceptions are marked in red. 

\subsection{Speed}
This section analyzes speed regulations. 
\begin{trafficrule}[Keep Control] 
	\blockquote{A person operating a vehicle may only travel at a speed that allows them to be in constant control of their vehicle} [§3(1) StVO].
\end{trafficrule}
Control is lost if vehicle tires cannot exert the required forces on the road. This happens when the lateral or longitudinal accelerations exceed the limits of the friction circle. Not to lose control is not only a rule but also a safety requirement for any autonomous vehicle. Hence control algorithms will limit the requested accelerations accordingly.

\begin{trafficrule}[Above Minimum Speed] 
	\blockquote{No motor vehicle must, without good reason, travel so slowly as to impede the flow of traffic} [§3(2) StVO; VC 13.4].
\end{trafficrule}

\citet{Decastro2018} define \blockquote{impeding the traffic flow} as going below a speed difference threshold $v_{\mathrm{ego}}-\overline{v}$. 
While they define $\overline{v}$ to be the average speed of the surrounding vehicles, it might be useful to damp this signal. 

\begin{trafficrule}[Below Speed Limit]
	Adhere to the "maximum permissible speed" [StVO 3(3)].
\end{trafficrule}
We assume the maximum speed limit to be available.
\begin{trafficrule}[No Stopping]
	On motorways and motor roads, \blockquote{stopping is prohibited, including on verges} [§18(8) StVO].
\end{trafficrule}

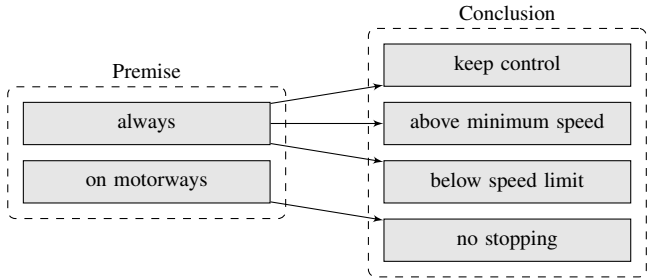
\begin{figure}[ht]
	\centering
	\begin{tikzpicture}
	\footnotesize
	\node [assumption] (a1) {always};
	\node [assumption, below=0.2cm of a1] (a2) {on motorways};
	\node [guarantee, right=1.5cm of a1] (g1) {above minimum speed};
	\node [guarantee, above=0.2cm of g1] (g2) {keep control};
	\node [guarantee, below=0.2cm of g1] (g3) {below speed limit};
	\node [guarantee, below=0.2cm of g3] (g4) {no stopping};
	
	\scoped[on background layer]
	\node [premise-conclusion, label=Premise, fit=(a1) (a2)] {};
	
	\scoped[on background layer]
	\node [premise-conclusion, label=Conclusion, fit=(g1) (g2) (g3) (g4)] {};
	
	\path [line] (a1) -- (g1);
	\path [line] (a1) -- (g2);
	\path [line] (a1) -- (g3);
	\path [line] (a2) -- (g4);
	\end{tikzpicture}
	\caption{Speed Limit Rules.}
	\label{fig:speed_limit}
	\vspace{-0.5cm}
\end{figure}

Motorways are defined as roads that are only allowed for motor vehicles, and they have specific entry and exit terminals \cite{EconomicCommissionforEuropeInlandTransportCommittee1968}. They usually consist of separate carriageways for two-way traffic. We do not consider motor roads in this study, as they have no dual carriageway. We assume the road type to be available.
\refFigure{fig:speed_limit} shows the codified speed limit rules. 

\subsection{Use of Roads and Lanes (Lane Selection)}
\label{subsec:lane_selection}
This section analyzes rules that specify which lanes should be used by motorists. 
\begin{trafficrule}[Keep Right]
	\blockquote{Keep as far to the right as possible} [§2(2) StVO; VC 10.3].
\end{trafficrule}

This is often referred to as \blockquote{staying on the right}.
However, in dense traffic, drivers might ignore the keep-right directive:
\begin{trafficrule}[\textit{Except.} Dense Traffic with Multiple Lanes]
	\blockquote{This might be ignored on carriageways with several lanes for one direction, [...] if traffic density justifies} [§7(1) StVO].
\end{trafficrule}
\citet{Wuthishuwong2013} define traffic density as the number of vehicles per lane that measure to the \SI{1}{\kilo\meter} length of the observed street. However, there are currently no traffic density values that autonomous vehicles can use. 

Another exception is made within built-up areas on roads that are no motorways:
\begin{trafficrule}[\textit{Except.} Built-up Area]
	\blockquote{On carriageways with several marked lanes for one direction of traffic [...] within built-up areas -- with the exception of motorways -- [...], vehicles [...] are free to choose their lane, even at no dense traffic} [§7(3) StVO].
\end{trafficrule}

\begin{figure}[ht]
	\centering
	\begin{tikzpicture}
	\footnotesize
	\node [assumption] (a1) {always};
	\node [assumption-except, below=0.2cm of a1] (a2) {not dense traffic};
	\node [assumption-except, below=0.2cm of a2] (a3) {not (in build-up area and not motorway)};
	\node [assumption-except, below=0.2cm of a3] (a4) {in build-up area or \#lanes $\leq$ 2};
	\node [assumption, below=0.2cm of a4] (a5) {not in build-up area and \#lanes $>$ 2};
	\node [guarantee, right=1.5cm of a3] (g1) {keep in the right-most lane};
	\node [guarantee, below=0.2cm of g1] (g2) {keep outside the left-most lane};
	
	\scoped[on background layer]
	\node [premise-conclusion, label=Premise, fit=(a1) (a5)] {};
	
	\scoped[on background layer]
	\node [premise-conclusion, label=Conclusion, fit=(g1) (g2)] {};
	
	\path [line] (a1) -- (g1);
	\path [line-except] (a2) -- (g1);
	\path [line-except] (a3) -- (g1);
	\path [line-except] (a4) -- (g1);
	\path [line] (a5) -- (g2);
	\end{tikzpicture}
	\caption{Lane Selection Rules.}
	\label{fig:rule_lane_selection}
	\vspace{-0.5cm}
\end{figure}
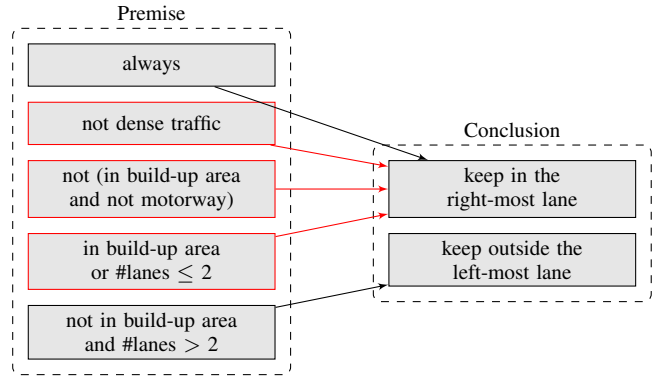

An exception to the keep right directive exists for outside built-up areas on roads with more than two lanes for one-way traffic. \blockquote{Inside a built-up area} is a synonym for inner-city.
\begin{trafficrule}[\textit{Except.} Outside Built-up Area With Three Or More Lanes]
	\blockquote{Outside built-up areas [with] three lanes for one direction of traffic, vehicles may, in derogation from the rule that they must keep as far to the right as possible, [stay in] the middle lane in places where -- even if only now and then -- a vehicle is stationary or moving in the nearside lane. On carriageways with more than three lanes marked in this way for one direction of traffic, the same applies to the second lane from the right} [§7(3c) StVO].
\end{trafficrule}
However, this exception brings in a new rule. We summarize the new rule as \blockquote{keep outside the left-most lane} and use the non-negated premise from the exception. 
\refFigure{fig:rule_lane_selection} shows the codified lane selection rules.

\subsection{Overtaking}
\label{subsec:overtaking}
The legal texts of the StVO and the Vienna Convention on Road Traffic are missing an explicit definition for overtaking. Whereas \citet{Rizaldi2017} define overtaking as the process of changing lane, passing a vehicle, and returning to your initial lane, court rulings have clarified that passing a vehicle is already considered as overtaking \cite{BGH4.Strafsenat1968a}. We will follow this definition and replace overtaking with passing whenever necessary. 

\begin{trafficrule}[Speed Advantage during Overtaking]
	Only overtake if the ego vehicle travels \blockquote{at a speed substantially higher than that of the vehicle to be overtaken} [§5(2) StVO].
\end{trafficrule}

As the term \blockquote{substantially higher speed} is vague, court rulings have since clarified the minimal speed advantage for trucks to be \SI{10}{\kilo\meter\per\hour} \cite{UberholenZweibrucken2009}. Missing other concrete values, we will use this value for passenger vehicles as well.

\begin{trafficrule}[Overtaking Maneuver]
	\blockquote{Make sure that traffic approaching from behind is not endangered. [Keep] a sufficient lateral distance [...] from other road users [...]. Move back to the right-hand side of the road as soon as possible} [§5(4) StVO, VC 11.2a; VC 11.4].
\end{trafficrule}

This rule can be divided into three parts. First, when changing to the outer lane, traffic shall not be jeopardized. \citet{Rizaldi2017} define this as keeping a safe distance to the new follower. Second, sufficient lateral space will be inherently satisfied by a motion planner. Thus, we will not consider it in this paper explicitly. Third, a vehicle shall move back as soon as possible. Following \citet{Rizaldi2017}, the phrase \blockquote{as soon} means when a safe distance to the new follower can be established. However, as stated before in \refSection{subsec:lane_selection}, there are multiple exceptions to this rule. We relax this rule in this work and interpret it as \blockquote{do not return to the initial lane before a safe distance can be established}. This means that when performing a lane-change, a safe distance to the rear vehicle should be ensured.

\begin{trafficrule}[No Right Overtaking]
	Only \blockquote{overtake [...] on the left} [§5(1) StVO; VC 11.1].
\end{trafficrule}

We will interpret this as passing a vehicle. This rule then also implies that vehicles on the right lane should not travel faster than those on the left. 
However, there are several exceptions for special lane types:
\begin{trafficrule}[\textit{Except.} Diverging Lane]
	\blockquote{Where lanes diverge from the main carriageway [...] vehicles turning off may [...] travel faster than traffic on the main carriageway} [§7a(1) StVO].
\end{trafficrule}

\begin{trafficrule}[\textit{Except.} Acceleration Lane]
	\blockquote{On motorways and other roads outside built-up areas, vehicles may travel faster in acceleration lanes than traffic on the main carriageway} [§7a(2) StVO].
\end{trafficrule}

\begin{trafficrule}[\textit{Except.} Deceleration Lane]
	\blockquote{If traffic on the main carriageway is moving slowly or is stationary, vehicles in a deceleration lane may overtake at a moderate speed} [§7a(3) StVO].
\end{trafficrule}

Another exception for built-up areas:
\begin{trafficrule}[\textit{Except.} Build-up Area]
	\blockquote{On carriageways with several marked lanes for one direction of traffic [...] within built-up areas -- with the exception of motorways -- [...], traffic on the right may move faster than traffic on the left} [§7(3) StVO].
\end{trafficrule}

Also, there are exceptions for dense traffic:
\begin{trafficrule}[\textit{Except.} Dense Traffic]
	In dense traffic, \blockquote{traffic on the right (nearside lane, middle lane) may move faster than traffic on the left} [§7(2) StVO].
\end{trafficrule}

\begin{trafficrule}[\textit{Except.} Dense Traffic]
	Vehicles queues at low speed or standstill may be overtaken \blockquote{on the right} [§7(2a) StVO; VC 11.6].
\end{trafficrule}

We argue that the above two exceptions essentially have the same meaning if overtaking is interpreted as passing: In dense traffic, passing vehicles on the left side is allowed.

\begin{figure}[ht]
	\centering
	\begin{tikzpicture}
	\footnotesize
	\node [assumption] (a1) {always};
	\node [assumption-except, below=0.2cm of a1] (a2) {not on diverging lane};
	\node [assumption-except, below=0.2cm of a2] (a3) {not on acceleration lane};
	\node [assumption-except, below=0.2cm of a3] (a4) {not dense traffic};
	\node [assumption-except, below=0.2cm of a4] (a5) {not (in built-up area and not motorway) };
	
	\node [assumption, below=0.2cm of a5] (a6) {not at substiantially higher speed};
	
	\node [guarantee, right=1.5cm of a1] (g1) {safe distance to rear vehicle during lane change};
	\node [guarantee, right=1.5cm of a3] (g2) {no right passing};
	\node [guarantee, right=1.5cm of a6] (g3) {no left passing};
	
	\scoped[on background layer]
	\node [premise-conclusion, label=Premise, fit=(a1) (a6)] {};
	
	\scoped[on background layer]
	\node [premise-conclusion, label=Conclusion, fit=(g1) (g3)] {};
	
	\path [line] (a1) -- (g1);
	\path [line] (a1) -- (g2);
	\path [line-except] (a2) -- (g2);
	\path [line-except] (a3) -- (g2);
	\path [line-except] (a4) -- (g2);
	\path [line-except] (a5) -- (g2);
	\path [line] (a6) -- (g3);
	\end{tikzpicture}
	\caption{Overtaking Rules.}
	\label{fig:overtaking}
	\vspace{-0.5cm}
\end{figure}
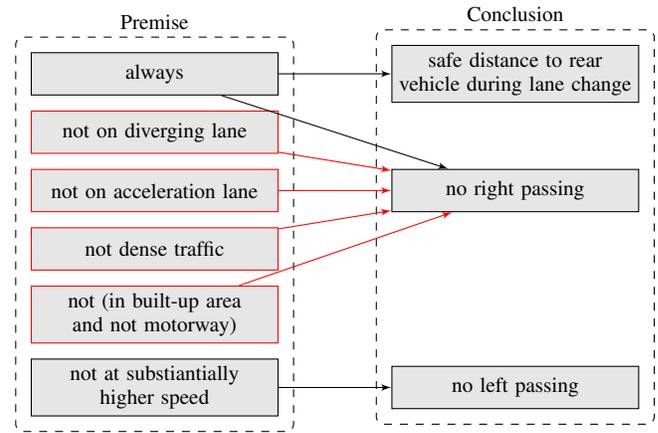

\refFigure{fig:overtaking} shows the codified overtaking rules.

\subsection{Safe Distance}
\label{subsec:safe_distance}
A driver shall always keep a safe distance to a preceding vehicle:
\begin{trafficrule}[Safe Distance]
	\blockquote{A person operating a vehicle moving behind another vehicle must, as a rule, keep a sufficient distance from that other vehicle so as to be able to pull up safely even if it suddenly slows down or stops} [§4(1) StVO; VC 13.5].
\end{trafficrule}

\begin{figure}[ht]
	\centering
	\begin{tikzpicture}
	\footnotesize
	\node [assumption] (a1) {always};
	
	\node [guarantee, right=1.5cm of a1] (g1) {safe distance to preceding vehicle};
	
	\scoped[on background layer]
	\node [premise-conclusion, label=Premise, fit=(a1)] {};
	
	\scoped[on background layer]
	\node [premise-conclusion, label=Conclusion, fit=(g1)] {};
	
	\path [line] (a1) -- (g1);
	\end{tikzpicture}
	\caption{Safe Distance Rule.}
	\label{fig:distance}
	\vspace{-0.5cm}
\end{figure}
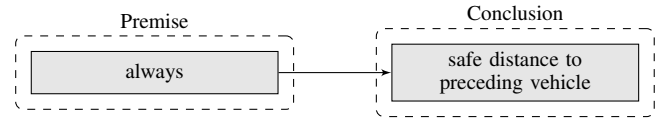

This rule has been treated frequently in literature \cite{Vanholme2013, ReyesCastro2013, Rizaldi2016, Shalev-Shwartz2017}, although implementations vary in the order of state derivatives used to calculate the distance. Estimating the maximum possible braking deceleration is also not clearly defined and may change depending on the road surface. \refFigure{fig:distance} shows the codified distance rule.

\subsection{Being Overtaken}
\label{subsec:being_overtaken}
A vehicle being overtaken shall obey to the following:
\begin{trafficrule}[Being Overtaken]
	A vehicle \blockquote{being overtaken must not increase the vehicle’s speed} [§5(6) StVO].
\end{trafficrule}

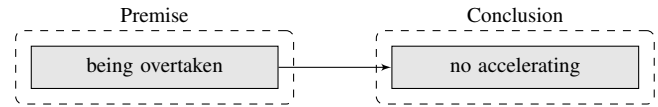
\begin{figure}[ht]
	\centering
	\begin{tikzpicture}
	\footnotesize
	\node [assumption] (a1) {being overtaken};
	
	\node [guarantee, right=1.5cm of a1] (g1) {no accelerating};
	
	\scoped[on background layer]
	\node [premise-conclusion, label=Premise, fit=(a1)] {};
	
	\scoped[on background layer]
	\node [premise-conclusion, label=Conclusion, fit=(g1)] {};
	
	\path [line] (a1) -- (g1);
	\end{tikzpicture}
	\caption{Being Overtaken Rule.}
	\label{fig:overtaken}
	\vspace{-0.5cm}
\end{figure}

We define being overtaken to be close to a vehicle on the left lane, which is similar to the definition used in \cite{Decastro2018}. This is sufficient, as overtaking on the right is prohibited.
\refFigure{fig:overtaken} shows the codified rule when being overtaken.

\subsection{Priorities}
\label{subsec:priorities}
This section deals with priority rules.
Giving way means that a driver \blockquote{continues or resumes his advance or maneuver if by so doing he might compel the drivers of other vehicles to change the direction or speed of their vehicle abruptly} \cite{EconomicCommissionforEuropeInlandTransportCommittee1968}. If a vehicle has the right of way, the other drivers shall be giving way. 
\begin{trafficrule}[Right of way]
	On \blockquote{motorways and motor roads [...] traffic on the main carriageway has the right of way} [§18(3) StVO; VC 25.2].
\end{trafficrule}

\begin{trafficrule}[Zipper Merge]
	\blockquote{If, on roads with several lanes for one direction, uninterrupted travel in one of the lanes is not possible, or if a lane comes to an end, vehicles traveling in the adjacent lane must allow vehicles in the other lane to change lanes immediately before the road narrows, in such a way as to let them join their line of traffic in turn after each vehicle traveling in the uninterrupted lane} [§7(4) StVO].
\end{trafficrule}
The rule demands that vehicles should merge at the end of the lane in an alternating zip fashion from both lanes. Following the zipper merge has proved to reduce congestion while ensuring the safety of motorists, as the complexity in changing lanes is removed. Some US states have begun adopting this concept \cite{Marshall2016}.

\begin{figure}[ht]
	\centering
	\begin{tikzpicture}
	\footnotesize
	\node [assumption] (a1) {motorways};
	\node [assumption, below=0.2cm of a1] (a2) {\#lanes$>$1 and travel flow in one lane interrupted};
	\node [assumption, below=0.2cm of a2] (a3) {lane ends};
	
	\node [guarantee, right=1.5cm of a1] (g1) {right of way};
	\node [guarantee, right=1.5cm of a2] (g2) {zipper merge};
	
	\scoped[on background layer]
	\node [premise-conclusion, label=Premise, fit=(a1) (a3)] {};
	
	\scoped[on background layer]
	\node [premise-conclusion, label=Conclusion, fit=(g1) (g2)] {};
	
	\path [line] (a1) -- (g1);
	\path [line] (a2) -- (g2);
	\path [line] (a3) -- (g2);
	\end{tikzpicture}
	\caption{Priorities Rules.}
	\label{fig:priorities}
	\vspace{-0.3cm}
\end{figure}
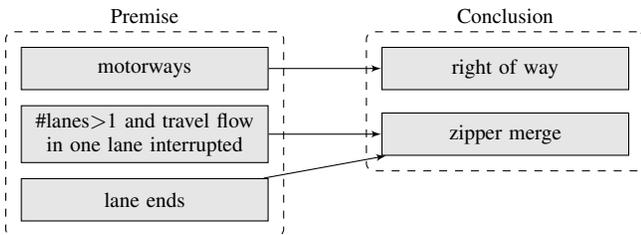

If a driver in a continuing lane does not obey the zipper merge, a driver wishing to merge is not allowed to enforce it \cite{ZipMergeAgDD2006}.
Thus, if an accident occurs during the zipper merge, the blame is often shared \cite{ZipMergeAgDD2006, ZipMergeMuenchen2008}.
Right of way and zipper merge are dual and thus contradicting, as StVO does not specify the application of the zipper merge to be an exception to the right of way rule. Future updates to the regulation should clarify this.
\refFigure{fig:priorities} shows the codified priority rules.

\section{Related Work}
\label{sec:related_work}
Works to formalize traffic rules have come from three different communities: First, the planning community, which tries to develop a planner that can follow all applicable rules. Here, the rules are checked on potential predicted outcomes \cite{Vanholme2013, ReyesCastro2013, Esterle2019a}. Second, the safety community, which has tried to establish contracts consisting of a set of rules, which every vehicle should adhere to prove safety \cite{Decastro2018, Shalev-Shwartz2017}. These approaches generally rely on inter-vehicle communication. Third, the legal community, which tries to analyze recorded traces to identify liability, which is relevant to insurance companies \cite{Rizaldi2015, Rizaldi2016, Rizaldi2017}. We will now discuss related works in detail.

\citet{Vanholme2013} were the first to perform a detailed analysis of the applicable rules for highway driving based on the Vienna Convention on Road Traffic. They used inequality comparisons of real numbers to express the rules. However, they did not provide a concrete formalization for most of the behavioral rules, such as overtaking or right of way. Also, since the zipper merge is not part of the Vienna Convention on Road Traffic, the authors did not elaborate on it either. \citet{ReyesCastro2013} used LTL to express traffic rules. Since the authors focused on developing a planning algorithm, they only provided examples for formalized rules such as \blockquote{do not cross solid center lines} or \blockquote{do not travel in the wrong direction}, which only depend on the ego vehicle itself.

\citet{Decastro2018} formalized the rules as contracts between vehicles. They used inequality comparison of real numbers to formalize the rules regarding lane selection, overtaking (safe, on the left), etc. However, they did not consider rules such as \blockquote{right of way} or \blockquote{zipper merge}, which are challenging because of the behavioral uncertainty of the agents involved. 
The \textit{Responsibility-Sensitivity Safety Model} (RSS) defined by \citet{Shalev-Shwartz2017} has formalized the notion of a safe distance and the right of way. 
Their approach can only check rules between an ego agent and one other agent \cite{Gassmann2019}.

\citet{Rizaldi2015} used higher-order logic to formalize parts of the Vienna Convention on Road Traffic. They extended their work in \cite{Rizaldi2016} to prove the correctness of the \blockquote{safe distance} predicate from the StVO using a theorem prover. In \cite{Rizaldi2017}, they used the \blockquote{safe distance} predicate for the safe overtaking rule from the StVO, formalized in LTL.

Others have used Signal Temporal Logic (STL) to obtain quantitative semantics about rule satisfaction \cite{Cho2019, Arechiga2019}. Quantitive semantics might be beneficial for relaxing the requirements to satisfy a rule. \citet{Cho2019} formalize basic rules concerning a safety envelope such as staying in lane or speed boundaries in Signal Temporal Logic, while \citet{Arechiga2019} formalize the safe distance from the RSS model in STL.

Lanelet2, a map framework for highly automated driving, provides an interface called \blockquote{regulatory elements} to retrieve traffic signs, traffic light, speed limits, and right of way \cite{Poggenhans2018}. For the \blockquote{right of way}, Lanelet2 provides the lanes on which vehicles have the right of way. More elaborate rules such as overtaking, distance keeping, or zipper merge are not included. 

\begin{table}[tb]
	\vspace{0.15cm}
	\footnotesize
	\centering
	\caption{Rule formulations in the literature. Categorization of technology being used into inequality comparison of real numbers (\techreal) and Linear Temporal Logic (\techltl). We use \playersinv{\textbullet} to state the number of other agents this rule depends on. If the rule is mentioned but a clear description of the implementation is missing, we use \techmissing. If a rule has not been discussed in the publication, we use --.}
	\begin{tabular}{L{3cm}L{0.80cm}L{0.80cm}L{0.8cm}L{1.14cm}}
		\toprule
		Rule & \citet{Vanholme2013} & \citet{Decastro2018} & \citet{Rizaldi2017} & \citet{Shalev-Shwartz2017}\\
		\midrule
		\textit{-- Lane Selection --} \\
		keep in right-most lane & \techmissing & \techreal \playersinv{0} & -- & -- \\
		keep outside left-most ln. & \techmissing & \techreal \playersinv{0} & -- & -- \\
		\midrule
		\textit{-- Overtaking --} \\
		safe lane change & \techmissing & \techreal \playersinv{2} & \techltl \playersinv{2} & -- \\
		no right passing & \techmissing & \techreal \playersinv{1} & -- & -- \\
		speed adv. during overtak. & -- & -- & -- & -- \\
		\midrule
		safe distance (preced.) & \techreal \playersinv{1} & \techreal \playersinv{1} & \techltl \playersinv{1} & \techreal \playersinv{1}\\
		\midrule
		being overtaken 	& -- & \techreal \playersinv{1} & -- & -- \\
		\midrule
		\textit{-- Priorities --} \\
		right of way 		& \techmissing & -- & -- & \techreal \playersinv{1} \\
		zipper merge 		& -- & -- & -- & -- \\
		\bottomrule
	\end{tabular}
	\label{tab:rule_literature}
	\vspace{-0.15cm}
\end{table}

\refTable{tab:rule_literature} summarizes our analysis, mapping the most relevant formalization works in literature to the identified rules from our legal analysis. We omit trivial rules e.g. \blockquote{no stopping} or special cases e.g. \blockquote{keep outside left-most lane}. Different techniques on how to model the rules have been employed. Formal logics such as LTL or STL, as well as real-value constraints, have been used. Presently, there is no comprehensive formalized set of traffic rules in literature.

\section{Formalizing Traffic Rules using Linear Temporal Logic}
Following our legal analysis, we will now formalize those rules in a formal language.
We will follow \citet{Rizaldi2017} to distinguish between codification (representing natural language specifications as logical entities) and concretization (concretely interpreting predicates).
\subsection{Linear Temporal Logic for Codification}
During the legal analysis, conjunction, disjunction, negation, and implication proved to be powerful and useful tools for formalizing rules. As traffic rules such as overtaking consider temporal behaviors, we decided to use LTL.

Formally, the language $\LTLformula$ of LTL formulas is defined as
\begin{align*}
\LTLformula ::= &\, \AtomicProposition\, | \, \lnot \LTLformula \, | \, \LTLformula_1 \land \LTLformula_2 \, | \LTLformula_1 \lor \LTLformula_2  | \LTLformula_1 \ImpliesCustomized \LTLformula_2 \, | \, \Next \LTLformula \, | \\ 
&\, \LTLformula_1 \lUntil \LTLformula_2 \, | \, \Global \LTLformula \, | \, \Final \LTLformula,
\nonumber
\end{align*}
where $\AtomicProposition \in \AtomicPropositionSet$ denotes an \emph{atomic} proposition, $\lnot$ (resp. $\land$, $\lor$, $\ImpliesCustomized$) denote the Boolean operators ``not'', ``and'', ``or'' and ``implies'',
and $\Next$, (resp. $\lUntil$, $\Global$, $\Final$) denote the temporal operators ``next'', ``until'', ``globally'' (or ``always''), ``finally'' (or ``eventually'').
Refer to \cite{Baier2008} for definitions of the semantics.

We will separate the rules into premise and conclusion 
\begin{equation*}
	\LTLformula = \Global (\LTLformula^p \implies \LTLformula^c).
\end{equation*}
It allows rules to be divided into a premise about the current state of the environment, i.e. when a rule applies, and the legal behavior of the ego agent in that situation (conclusion). Then, exceptions to the rules can be modeled to be part of the assumption.

\subsection{Predicates for Concretization}
First, we need to identify suitable predicates, then, provide a function or formula to calculate them. For some labels, such as collision or speed limit violation, this is trivial. For others, such as the notion of safety, it is not. \citet{Rizaldi2016} thus used a theorem prover to prove the safe distance predicate.

Maps are currently an important part of automated driving. Map frameworks such as Lanelet2, provide location information (built-up vs. non-built-up) and road types (road, highway) \cite{Poggenhans2018}. We assume special lane types such as diverging or accelerating lanes to be available.

\begin{table}[tb]
	\renewcommand{\labelstyle}[1]{\scriptsize{\textsf{\textit{{#1}}}}}
	\vspace{0.15cm}
	\footnotesize
	\centering
	\caption{Atomic propositions and their respective interpretations from the perspective of agent  $\Agenti[\labelEgoAgent]$ in relation to another agent $\Agenti[\labelInRelationToAgent]$.}
	\begin{tabular}{L{1.6cm}L{6.0cm}}
		\toprule
		$\AtomicProposition \in \AtomicPropositionSet$ & Interpretation\\
		\midrule
		$\labelDense\ionly$ & $\Agenti[\labelEgoAgent]$ is closer than $\DenseRadius$ to $\DenseNumAgents$ or more agents \\
		$\labelInDirectFront\itoj$ & $\Agenti[\labelEgoAgent]$ is the predecessor of $\Agenti[\labelInRelationToAgent]$ \\
		$\labelRight\itoj$ & $\Agenti[\labelEgoAgent]$ is to the right of $\Agenti[\labelInRelationToAgent]$ \\
		$\labelLeft\itoj$ & $\Agenti[\labelEgoAgent]$ is to the left of $\Agenti[\labelInRelationToAgent]$ \\
		$\labelFront\itoj$ & $\Agenti[\labelEgoAgent]$ is in the front of $\Agenti[\labelInRelationToAgent]$ \\
		$\labelBehind\itoj$ & $\Agenti[\labelEgoAgent]$ is behind of $\Agenti[\labelInRelationToAgent]$ \\
		$\labelMerged\ionly$ & $\Agenti[\labelEgoAgent]$ has passed a static merging point, from which on a merge is not possible anymore \\
		$\labelSafeDistanceFront\ionly$ & $\Agenti[\labelEgoAgent]$ has a safe distance to the preceding vehicle \\
		$\labelSafeDistanceRear\ionly$ & $\Agenti[\labelEgoAgent]$ has a safe distance to the following vehicle \\
		$\labelColliding\ionly$ & $\Agenti[\labelEgoAgent]$ is colliding with road boundaries or any other agent or obstacle \\
		$\labelLaneChange\ionly$ & $\Agenti[\labelEgoAgent]$ is crossing a lane boundary\\
		$\labelNear\itoj$ & $\Agenti[\labelEgoAgent]$ is closer than $\NearDistance$ to $\Agenti[\labelInRelationToAgent]$ \\
		$\labelLaneEnd\ionly$ & $\Agenti[\labelEgoAgent]$ has less than $\RemainingDistance$ remaining to the end of the lane\\
		$\labelAccelerate\ionly$ & $\Agenti[\labelEgoAgent]$ accelerates with $a>\AccLimit$\\
		$\labelSpeedDiff\itoj$ & $\Agenti[\labelEgoAgent]$ is faster than $\Agenti[\labelInRelationToAgent]$ and some threshold  $\VelocityThreshold$\\
		$\labelBuiltup\ionly$ & $\Agenti[\labelEgoAgent]$ is within a built-up area\\
		$\labelMotorway\ionly$ & $\Agenti[\labelEgoAgent]$ is on a road type: motorway\\
		$\labelDivergingLane\ionly$ & $\Agenti[\labelEgoAgent]$ is on a lane type: diverging lane\\
		$\labelAccelerationLane\ionly$ & $\Agenti[\labelEgoAgent]$ is on a lane type: acceleration lane\\
		\bottomrule
	\end{tabular}
	\label{tab:rule_predicates}
	\vspace{-0.15cm}
\end{table}

\begin{figure}[tb]
\vspace{0.15cm}
\footnotesize
\renewcommand{\labelstyle}[1]{\scriptsize{\textsf{\textit{{#1}}}}}
\centering
\def\svgwidth{\columnwidth}
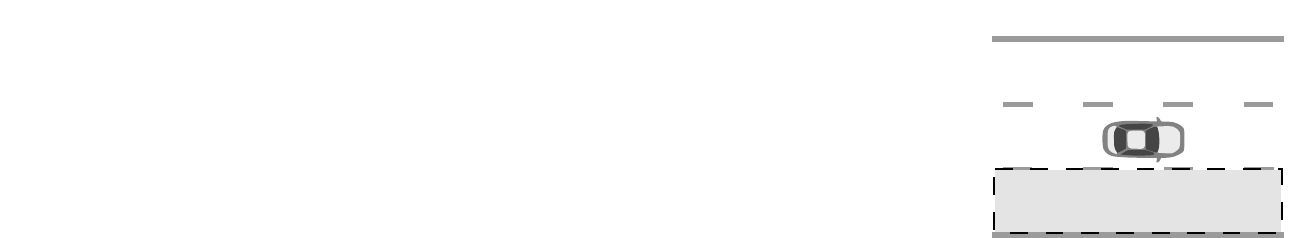
\caption{Relational labels, which describe that agent $\Agenti[\labelEgoAgent]$ located in the gray area is in $\labelFront/ \labelBehind/ \labelLeft/ \labelRight$ to agent $\Agenti[\labelInRelationToAgent]$.}
\label{fig:relational_labels}
\vspace{-0.3cm}
\end{figure}

\begin{table*}[tb]
	\renewcommand{\labelstyle}[1]{\scriptsize{\textsf{\textit{{#1}}}}}
	\vspace{0.15cm}
	\centering
	\caption{Formalized rules in the format $\LTLformula = \protect\Global (\LTLformula^p \implies \LTLformula^c)$.}
	\begin{tabular}{L{3.0cm}L{5.2cm}L{7.8cm}}
		\toprule
		Rule & Premise $\LTLformula^p$ & Conclusion $\LTLformula^c$ \\
		\midrule
		no right passing & $\lnot \labelDivergingLane\ionly \land \lnot \labelAccelerationLane\ionly \land \lnot \labelDense\ionly \land \lnot \labelBuiltup \lor \labelMotorway\ionly$  & $\lnot \big(\labelBehind\itoj \land \Next (\labelBehind\itoj \lUntil \labelRight\itoj \lUntil \labelFront\itoj)\big)$ \\
		safe lane change & $\labelLaneChange$ & $\labelSafeDistanceRear$ \\
		speed advantage for overtaking & $\labelBehind\itoj \land \Next (\labelBehind\itoj \lUntil \labelLeft\itoj \lUntil \labelFront\itoj)$ & $\labelSpeedDiff\itoj \lUntil \labelFront\itoj$ \\
		\midrule
		safe distance (preced.) & $\true$ & $\labelSafeDistanceFront$ \\
		\midrule 
		being overtaken & $\labelRight\itoj \land \labelNear\itoj$ & $\lnot \labelAccelerate\ionly$ \\
		\midrule
		zipper merge & $\LTLformula_{\mathrm{zip-sit}} \land \labelInDirectFront\itoj \land \lnot \labelMerged\ionly 
		\land (\labelInDirectFront\itoj \lor \labelMerged^{(\labelInRelationToAgent)}) \lUntil \labelMerged\ionly $
		& $\Global(\labelMerged\ionly \land \labelMerged^{(\labelInRelationToAgent)} \ImpliesCustomized \lnot \labelInDirectFront\itoj)$ \\
		\bottomrule
	\end{tabular}
	\label{tab:formalized_rules}
	\vspace{-0.15cm}
\end{table*}

\refTable{tab:rule_predicates} shows the predicates. We evaluate them based on the observed scene (from simulation or dataset replay) before passing them to the rule monitor, which will evaluate the rule formula at each time step. The relational labels are calculated according to \refFigure{fig:relational_labels}, i.e. in a partially overlapping manner.

\subsection{Codified Rules}
\refTable{tab:formalized_rules} shows the formalized rules, which we will discuss in the following. We leave the codification of lane selection rules (\refSection{subsec:lane_selection}) and right of way (\refSection{subsec:priorities}) to future work.
\subsubsection{Overtaking}
Based on our previous legal analysis and the usage of De Morgan's laws, we formalize three rules for overtaking (see \refTable{tab:formalized_rules}). Both \blockquote{no right passing} and \blockquote{speed advantage during overtaking} use the passing sequence behind -- left -- front. Note that in contrast to our previous work \cite{Esterle2019a}, we define the relational labels as partially overlapping, which changes the meaning of the rule from overtaking to passing.
The third rule \blockquote{safe distance at each lane change} covers the rules in \cite{Rizaldi2017}, where the authors defined overtaking as the process of changing lanes and passing, for which they imposed a safe rear distance at the beginning and finishing.

\subsubsection{Distance}
We formalize the rule to ensure a safe distance to any preceding vehicle, based on the calculation of the safe distance predicate in \cite{Rizaldi2017}.

\subsubsection{Being Overtaken}
We define \blockquote{being overtaken} as to be on the right and close to a vehicle. In such a situation, we prohibit to accelerate.

\subsubsection{Zipper Merge}
\begin{figure}[tb]
	\footnotesize
	\centering
	\def\svgwidth{0.4\columnwidth}
\begingroup%
  \makeatletter%
  \providecommand\color[2][]{%
    \errmessage{(Inkscape) Color is used for the text in Inkscape, but the package 'color.sty' is not loaded}%
    \renewcommand\color[2][]{}%
  }%
  \providecommand\transparent[1]{%
    \errmessage{(Inkscape) Transparency is used (non-zero) for the text in Inkscape, but the package 'transparent.sty' is not loaded}%
    \renewcommand\transparent[1]{}%
  }%
  \providecommand\rotatebox[2]{#2}%
  \newcommand*\fsize{\dimexpr\f@size pt\relax}%
  \newcommand*\lineheight[1]{\fontsize{\fsize}{#1\fsize}\selectfont}%
  \ifx\svgwidth\undefined%
    \setlength{\unitlength}{160.82632711bp}%
    \ifx\svgscale\undefined%
      \relax%
    \else%
      \setlength{\unitlength}{\unitlength * \real{\svgscale}}%
    \fi%
  \else%
    \setlength{\unitlength}{\svgwidth}%
  \fi%
  \global\let\svgwidth\undefined%
  \global\let\svgscale\undefined%
  \makeatother%
  \begin{picture}(1,0.3520664)%
    \lineheight{1}%
    \setlength\tabcolsep{0pt}%
    \put(0,0){\includegraphics[width=\unitlength,page=1]{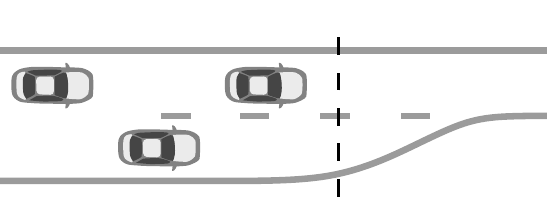}}%
    \put(-0.02298262,0.17174023){\color[rgb]{0,0,0}\makebox(0,0)[lt]{\lineheight{1.25}\smash{\begin{tabular}[t]{l}$\Agenti[\labelEgoAgent]$\end{tabular}}}}%
    \put(0.34534445,0.17174023){\color[rgb]{0,0,0}\makebox(0,0)[lt]{\lineheight{1.25}\smash{\begin{tabular}[t]{l}$\Agenti[\labelInRelationToAgent]$\end{tabular}}}}%
    \put(0.43052289,0.30482503){\color[rgb]{0,0,0}\makebox(0,0)[lt]{\lineheight{1.25}\smash{\begin{tabular}[t]{l}last merging point\end{tabular}}}}%
    \put(0,0){\includegraphics[width=\unitlength,page=2]{zipmerge.pdf}}%
    \put(0.1556476,0.06881936){\color[rgb]{0,0,0}\makebox(0,0)[lt]{\lineheight{1.25}\smash{\begin{tabular}[t]{l}$\Agenti[\labelInRelationToSecondAgent]$\end{tabular}}}}%
  \end{picture}%
\endgroup%

	\caption{Naming convention of $\Agenti[\labelEgoAgent]$, $\Agenti[\labelInRelationToAgent]$, $\Agenti[\labelInRelationToSecondAgent]$ during zipper merge.}
	\label{fig:zipmerge}
	\vspace{-0.4cm}
\end{figure}

We describe a zipping situation as 
\begin{equation*}
	\LTLformula_{\mathrm{zip-sit}} = \labelLeft^{(\labelEgoAgent \inrelationto \labelInRelationToSecondAgent)} \land \lnot\labelFront^{(\labelEgoAgent \inrelationto \labelInRelationToSecondAgent)} \land \labelNear^{(\labelEgoAgent \inrelationto \labelInRelationToSecondAgent)} \land \labelLaneEnd^{(\labelInRelationToSecondAgent)},
\end{equation*}
where an agent $\Agenti[\labelEgoAgent]$ is to the left of an agent $\Agenti[\labelInRelationToSecondAgent]$ in an ending or blocked lane, and is following another agent $\Agenti[\labelInRelationToAgent]$. \refFigure{fig:zipmerge} shows the naming conventions for this rule.
If the assumption in \refTable{tab:formalized_rules} is fulfilled, the rule should guarantee that agent $\Agenti[\labelInRelationToAgent]$ will not be directly in front of agent $\Agenti[\labelEgoAgent]$ after the merging point, as agent $\Agenti[\labelInRelationToSecondAgent]$ from the other lane has merged in-between and has become the new predecessor of agent $\Agenti[\labelEgoAgent]$. 

\section{Evaluation}
\label{sec:evaluation}
Evaluating the formalized rules on recorded drives of humans helps us to validate the rules, and, once the predicates are fixed, provides valuable insight in the extent to which humans follow the rules. We use the INTERACTION dataset \cite{Zhan2019}, which focuses on dense interactions, and analyze the compliance of each vehicle to the traffic rules. To the best of our knowledge, \cite{Rizaldi2016} forms the only work that evaluated their formalized traffic rule to do so. 

\subsection{Evaluation Methods and Dataset Processing}
We study our approach in the benchmarking framework BARK proposed in \cite{Bernhard2020}. We use Spot \cite{Duret-Lutz2016}, a C++ library for model checking, to translate the formalized LTL formula to a deterministic finite automaton, and to manipulate the automatons. Each rule is then captured in a {\classnamestyle{RuleMonitor}} object, which we use to monitor rule compliance throughout the simulation, effectively replaying the dataset. We analyze the two-way merge scenario {\classnamestyle{DR\_DEU\_Merging\_MT}} and the three-lane road lower part of the Chinese highway merging scenario {\classnamestyle{DR\_CHN\_Merging\_ZS}}. 

\subsection{Evaluation of Violation on Public Data}
\begin{figure}[tb]
	\vspace{0.15cm}
	\centering
	\input{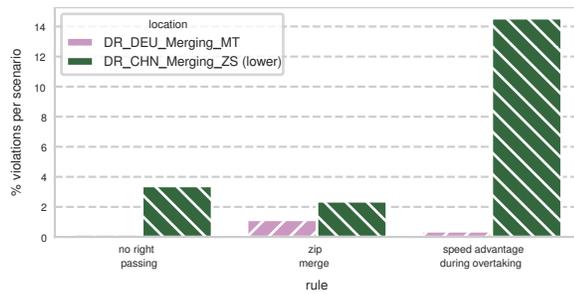}
	\caption{Once-per-agent violations of rules the dataset.}
	\label{fig:dataset_event_results}
	\vspace{-0.3cm}
\end{figure}

\begin{figure}[tb]
	\centering
	\input{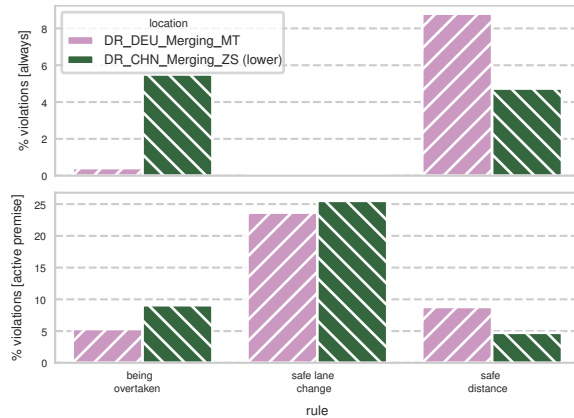}
	\caption{Violations per time of rules on the dataset.}
	\label{fig:dataset_time_results}
	\vspace{-0.5cm}
\end{figure}

For evaluation, we set $\AccLimit=\SI{0.5}{\meter\per\square\second}$,  $\VelocityThreshold=\SI{10}{\kilo\meter\per\hour}$, $\DenseRadius=\SI{20}{\meter}$ and $\DenseNumAgents=\num{8}$. We set $\NearDistance=\SI{5}{\meter}$ for the \blockquote{zipper merge} rule and $\NearDistance=\SI{3}{\meter}$ for the \blockquote{being overtaken} rule. For the Chinese data, we use $\RemainingDistance=\SI{20}{\meter}$. For the German data, we use $\RemainingDistance=\SI{55}{\meter}$, as the lane gets thinner much earlier. We use the parameters in \cite{Rizaldi2016} for the \blockquote{safe distance} label, and set the reaction time to $\SI{1}{\second}$. 
\paragraph{Once-per-agent violations}
We first study those rules, which premises contain temporal sequences, namely the \blockquote{no right passing}, \blockquote{zipper merge} and \blockquote{speed advantage during overtaking}. \refFigure{fig:dataset_event_results} shows the percentage of violation per agent. Once an agent violates a rule, that agent is flagged. About \num{1}\% of the vehicles in the German scenario violate the zipper merge. In the Chinese scenario, more than \num{2}\% do not merge according to the rule. The rule to have a significant \blockquote{speed advantage during overtaking} is much more often violated in the Chinese scenario than in the German. This could either stem from the distinct regulations, different local interpretation of what significant speed advantage means, or is due to the differences of the traffic situations. To study this, finding correlations between traffic features such as traffic density and the specific rule violations will be subject of future works.
\paragraph{Violations per time}
For the other rules, we study the relative number of violations per time, see \refFigure{fig:dataset_time_results}. We normalize the number of violations based on (1) the full dataset duration and (2) the duration when the premise was active. For the safe distance, the premise is always true, and thus the two violation metrics are identical. For others, the metric based on the premise being active is a more expressive violation metric. Violations close to \num{100}\% would probably indicate an error in the formalization. As the premise for \blockquote{safe lane change} by our definition is only active at one time instant, no significant violations can be observed when normalizing it by the complete simulation time. However, \refFigure{fig:dataset_time_results} shows that roughly every fourth lane change does not keep a safe distance to the rear vehicle, which is similar for the German and Chinese Data. We observe that in the German scenario, drivers do not keep a safe distance at about \num{8}\% of the time. In China, this value drops to about \num{4}\%. Note that we observe many more violations of the safe distance than in \cite{Rizaldi2016}, which can be explained by the fact that we include the reaction time from \cite{Rizaldi2017} in our calculation.

\section{Conclusion and Future Work}
\label{sec:conclusion}
We have formalized traffic rules for dual carriageways according to German traffic regulations. For this, we presented a methodology for legal analysis, which allowed us to codify these rules. We identified definition gaps in the regulations, especially in the predicates. We hope to start a discussion to remove this ambiguity. 
Our evaluation on real data helped us to concretize the predicates iteratively, and it also showed that humans violate formal traffic rules to a varying extent. We plan to extend our evaluation to more scenarios.
As a next step, false negatives and false positives need to be identified through a more elaborate evaluation, i.e. finding the correlations between rule violations. Also, the evaluation shall be extended to include lane selection rules and right of way.
Our work lays the foundation for integrating traffic rules into a planning component and leveraging the benefits of formalization to evaluate the rules' compliance.
Future work needs to collaborate with legal experts to verify our legal analysis and extend it to cover all regulations. 

\renewcommand{\bibfont}{\small}
\printbibliography
\end{document}